\documentclass{article}
\usepackage{ijcai21}

\usepackage{times}
\usepackage{soul}
\usepackage{url}
\usepackage[hidelinks]{hyperref}
\usepackage[utf8]{inputenc}
\usepackage[small]{caption}
\usepackage{graphicx}
\usepackage{amsmath}
\usepackage{amsthm}
\usepackage{booktabs}
\usepackage{algorithm}
\usepackage{algorithmic}
\usepackage{epsfig}
\usepackage{multirow}
\usepackage{bm}
\usepackage[table,xcdraw]{xcolor}
\usepackage{verbatim}
\usepackage{color}
\usepackage{arydshln}
\usepackage{amsthm,amsmath,amssymb}
\usepackage{mathrsfs}

\definecolor{mygray}{gray}{.88}
\definecolor{mygrayd}{gray}{.94}

\definecolor{citecolor}{RGB}{119,185,0}
\definecolor{citecolor1}{RGB}{66,168,235}
\definecolor{linkcolor}{RGB}{255,0,0}
\definecolor{urlcolor}{RGB}{255,105,180}
\usepackage{hyperref}
\hypersetup{colorlinks=true,citecolor=citecolor1,linkcolor=linkcolor,urlcolor=urlcolor}

\title{One-Shot Affordance Detection}

\author{
Hongchen Luo$^{1}$ \and
Wei Zhai$^{1,3}$\thanks{Wei Zhai is an intern at JD Explore Academy.} \and
Jing Zhang$^{2}$\thanks{Corresponding Author} \and
Yang Cao$^{1}$\footnotemark[2] \and
Dacheng Tao$^{3}$
\affiliations
\noindent
$^{1}$ University of Science and Technology of China, China \qquad \\
$^{2}$ The University of Sydney, Australia \qquad \\
$^{3}$ JD Explore Academy, JD.com, China \qquad \\
\emails
\{lhc12, wzhai056\}@mail.ustc.edu.cn,
jing.zhang1@sydney.edu.au,
forrest@ustc.edu.cn,
dacheng.tao@gmail.com
}

\begin{document}

\maketitle

\begin{abstract}

Affordance detection refers to identifying the potential action possibilities of objects in an image, which is an important ability for robot perception and manipulation. To empower robots with this ability in unseen scenarios, we consider the challenging \textbf{one-shot affordance detection} problem in this paper, i.e., given a support image that depicts the action purpose, all objects in a scene with the common affordance should be detected. To this end, we devise a One-Shot Affordance Detection (OS-AD) network that firstly estimates the purpose and then transfers it to help detect the common affordance from all candidate images. Through collaboration learning, OS-AD can capture the common characteristics between objects having the same underlying affordance and learn a good adaptation capability for perceiving unseen affordances. Besides, we build a Purpose-driven Affordance Dataset (PAD) by collecting and labeling 4k images from 31 affordance and 72 object categories. Experimental results demonstrate the superiority of our model over previous representative ones in terms of both objective metrics and visual quality. The benchmark suite is at  \href{https://lhc1224.github.io/lhc.github.io/}{ProjectPage}.

\end{abstract}

\begin{figure}[t]
	\centering
		\includegraphics[width=8.2cm]{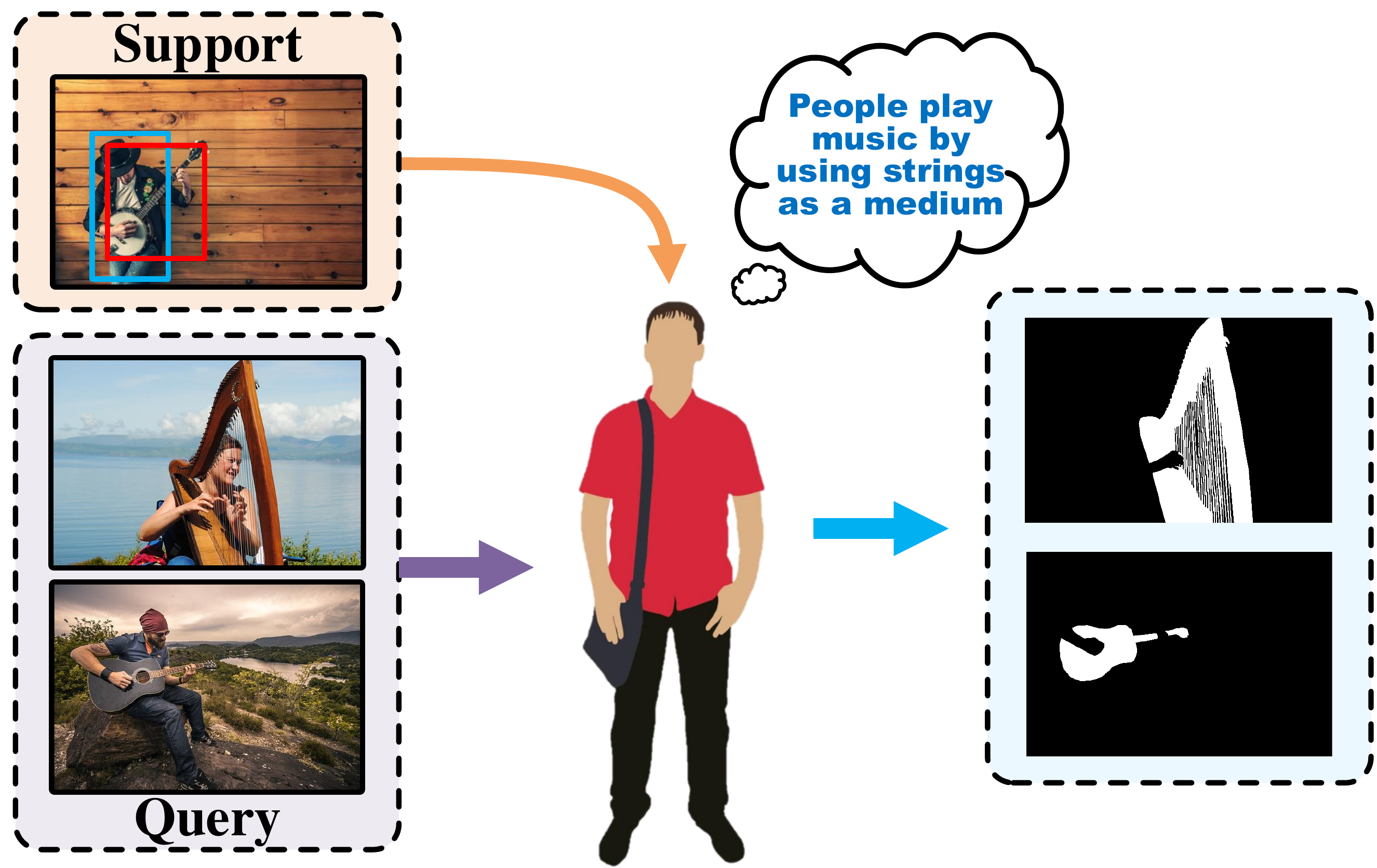}
	\caption{Illustration of perceiving affordance. Given a support image that depicts the action purpose, all objects in a scene with the common affordance could be detected.}
	\label{FIG:1}
\end{figure}

\section{Introduction}
The concept of affordance was proposed by the ecological psychologist Gibson \cite{gibson1977theory}. It describes how the inherent ``value'' and ``meanings'' of objects in an environment are directly perceived, and explains how this information can be linked to the action possibilities offered to an organism by the environment \cite{hassanin2018visual}. After Gibson put forward the definition of affordance, a lot of cognitive psychologists made profound studies on the relationship between object affordance and perceiver intention in recent decades \cite{heft1989affordances}. In particular, perceiving object affordance in unseen scenarios is a valuable capability and has a wide range of applications in scene understanding, action recognition, robot manipulation, and Human-Computer Interaction \cite{zhang2020empowering}.

\par
\begin{figure*}[t]
	\centering
		\includegraphics[scale=.285]{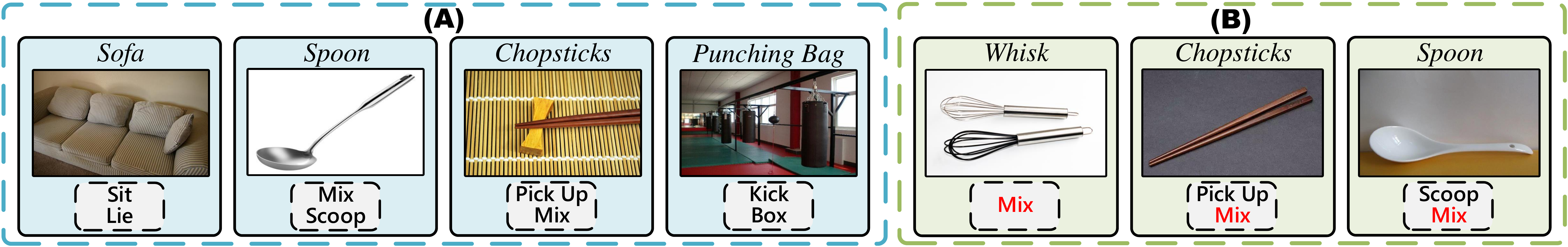}
	\caption{The part (A) shows that objects usually have multiple affordance. The part (B) shows that the objects with different semantic categories may have the same affordance.}
	\label{FIG:1_2}
\end{figure*}

 To learn such capability of perceiving affordance, we consider the challenging \textbf{one-shot affordance detection $\footnote[1]{``Detection'' follows the term pixel-wise detection task, which has also been used in the area of salient object detection.}$} problem in this paper, i.e., given a support image that depicts the action purpose, all objects in a scene with the common affordance should be detected (see Figure~\ref{FIG:1}). Unlike the object detection/segmentation problem \cite{shaban2017one}, affordance and semantic categories of objects are highly inter-correlated but do not imply each other. An object may have multiple affordances (see Figure~\ref{FIG:1_2}), e.g., the sofa can be used to sit or lie down. Actually, the possible affordance depends on the person's purpose in real-world application scenarios. Directly learning the affordance from a single image without the guidance of purpose makes the model tend to focus on the statistically dominant affordances while ignoring other visual affordances that are possibly suitable for completing the task.

\par To address this problem: 1) We try to find clear hints about the action purpose (i.e., via the subject and object locations \cite{chen2020recursive}) from a single support image, which implicitly defines the object affordance and is a reasonable setting in real-world unseen scenarios. 2) We adopt collaboration learning to capture the inherent relationship between different objects to counteract the interference caused by visual appearance differences and to improve generalization. Specifically, we devise a novel \textbf{O}ne-\textbf{S}hot \textbf{A}ffordance \textbf{D}etection (\textbf{OS-AD}) network to solve the problem. Taking an image as support and a set of images (5 images in this paper) as a query, the network first captures the human-object interactions from the support image using a purpose learning module (PLM) to encode the action purpose. Then, a purpose transfer module (PTM) is devised to use the encoding of the action purpose to activate the features in query images that have the common affordance. Finally, a collaboration enhancement module (CEM) is devised to capture the intrinsic relationships between objects with the same affordance and suppress backgrounds that are irrelevant to the action purpose. In this way, OS-AD can learn a good adaptation capability for perceiving unseen affordances.

\par Moreover, the existing datasets still have gaps relative to real application scenarios, due to the limitation of its diversity. The affordance detection for scene understanding and general applications should be able to learn from the human-object interaction when the robot arrives at a new environment and retrieves the objects in the environment, rather than just finding objects with the same categories or similar appearance. To address the limitations of the datasets, we propose the \textbf{P}urpose-driven \textbf{A}ffordance \textbf{D}ataset (\textbf{PAD}), which contains 4,002 diverse images covering 31 affordance categories as well as 72 object categories from different scenes. Moreover, we trained several representative models to address the problem and compare them with our OS-AD model comprehensively in terms of both objective evaluation metrics and visual quality. 

\noindent\textbf{Contributions }
(1) We introduce a new one-shot affordance detection problem along with a benchmark to facilitate the research for empowering robots with the ability to perceive unseen affordances in real-world scenarios.
(2) We propose a novel OS-AD network that can efficiently learn the action purpose and use it to detect the common affordance of all objects in a scene via collaboration learning, resulting in a good adaptation capability that can deal with unseen affordances.
(3) Experiments on the proposed PAD benchmark demonstrate that OS-AD outperforms state-of-the-art models and can serve as a strong baseline for future research.

\begin{figure*}[t]
	\centering
		\includegraphics[scale=.092]{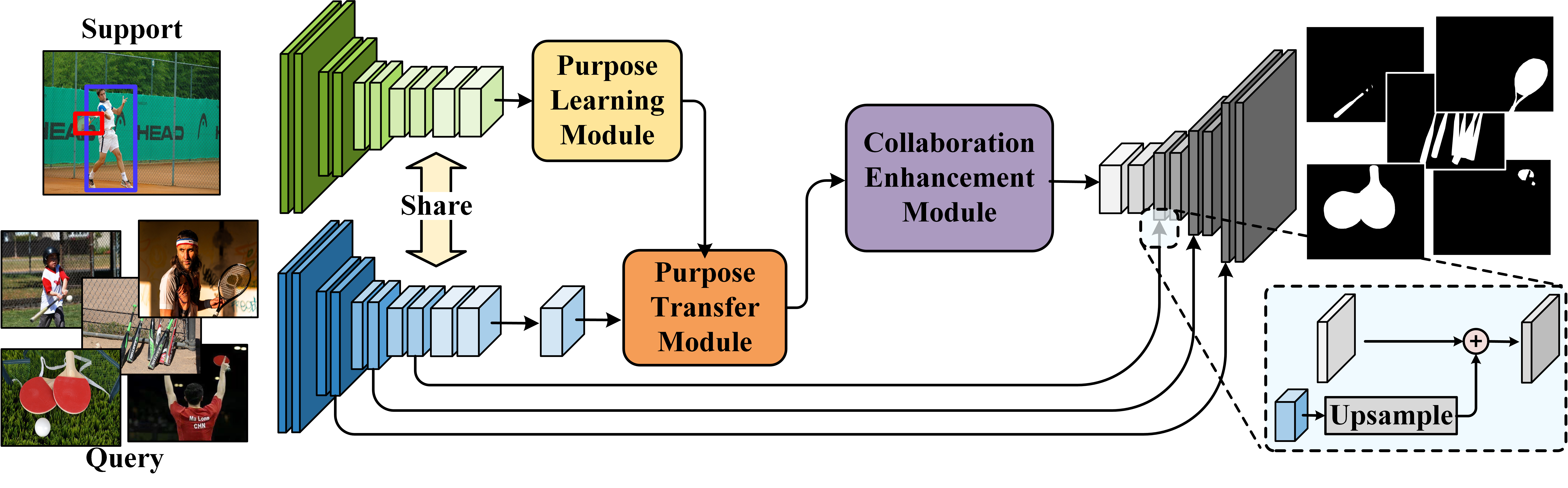}
	\caption{Our One-Shot Affordance Detection (OS-AD) network. OS-AD consists of three key modules: Purpose Learning Module (PLM), Purpose Transfer Module (PTM), and Collaboration Enhancement Module (CEM), which are detailed in Figure~\ref{FIG:4}.}
	\label{FIG:3}
\end{figure*}

\section{Related Work}

\subsection{Affordance Detection}
The problem of visual affordance detection has been investigated for decades \cite{hassanin2018visual}, it is a widely used strategy in AI community to perceive action intentions and thereby to infer visual affordance from the image/video of human and objects. Early works mainly attempted to establish and learn an association between the apparent characteristics of objects and their affordance for perceiving affordance. \cite{myers2015affordance} proposed a framework for jointly locating and identifying the affordance of object parts and presented the RGB-D Part Affordance dataset, which is the first pixel-wise labeled affordance dataset. However, the affordances of objects do not simply correspond to representational characteristics, which are shifted in response to the state of interactions between objects and humans. Therefore, \cite{chuang2018learning} considered the problem of affordance reasoning in the real world by taking into account both the physical world and the social norms imposed by the society, and constructed the ADE-Affordance dataset based on ADE20k \cite{zhou2017scene}. Since the change of an object's affordance state is usually due to the interaction between the human and the object, further research has begun to consider human action as a cue for learning affordance. \cite{demo2vec2018cvpr} used human-object interactions in demonstration videos to predict the affordance regions of static objects via linking human actions to object affordance, and proposed the OPRA dataset for affordance reasoning. Different from the above existing works, our proposed method aims to explore the multiplicity of affordance by a collaborative learning strategy. To a certain extent, our work conforms to Gibson’s definition of affordance that “it implies the complementarity of the animal and the environment”. Since there exist multiple potential complementarities between animal and environment, it leads to multiple possibilities of particular affordance. To address this issue, we present a novel task of one-shot affordance detection, in which action intension is introduced through support image to alleviate the multiplicity of affordance.

\subsection{One-Shot Learning}

Recently, one-shot learning has received a lot of attention and substantial progress has been made based on metric learning using the siamese neural network \cite{he2020progressive,koch2015siamese,snell2017prototypical}. Besides, some works build upon meta-learning and generation models to achieve one-shot learning. Specifically, \cite{michaelis2018one} proposed the problem of one-shot segmentation in clutter, finding and segmenting a previously unseen object in a cluttered scene based on a single instruction example. \cite{zhu2019one} proposed a one-shot texture retrieval, given an example of a new reference texture, detecting and segmenting all the pixels of the same texture category within an arbitrary image. Different from these studies, we aim to identify the potential action possibilities (affordance) of objects in unseen scenarios, given a single support image that implicitly defines the action purpose without an explicit mask to denote the object affordance, which is very challenging but of practical value.

\section{Method}
Our One-Shot Affordance Detection (OS-AD) network is shown in Figure~\ref{FIG:3}. During training, it receives a single support image that implicitly defines the affordance via human-object interactions, e.g., in the form of bounding boxes of the person and the object interacting with the person, which are weaker signals and easier to acquire than a pixel-wise affordance mask. Meanwhile, a set of query images (5 in this paper) containing objects with the same affordance. Our goal is to estimate the action purpose from the support image, transfer it to the query images, and learn to segment all the objects having the same affordance. To this end, three modules named PLM, PTM, and CEM are devised. During testing, OS-AD can recognize the common affordance from any number of query images given a single support image. 

\begin{figure*}[t]
	\centering
		\includegraphics[scale=.094]{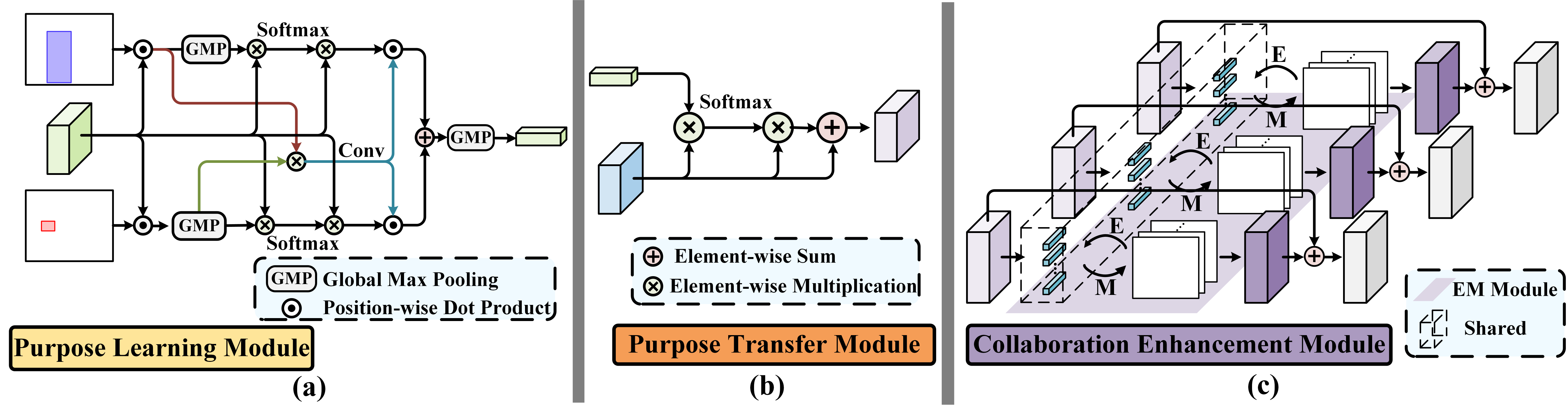}
	\caption{(a) PLM aims to estimate action purpose from the human-object interaction in the support image. (b) PTM transfers the action purpose to the query images via an attention mechanism to enhance the relevant features. (c) CEM captures the intrinsic characteristics between objects having the common affordance to learn a better affordance perceiving ability.}
	\label{FIG:4}
\end{figure*}

\subsection{Framework}

Given a set of query images $\mathcal{I}=\{I_1,...,I_n\}$ and a support image $I_{\text{sup}}$ containing human-object interactions, we first extract the features of $\mathcal{I}$ and $I_{\text{sup}}$ using resnet50 \cite{he2016deep} to obtain their feature representations $\mathcal{X}=\{X_1,...,X_n\}$ and $X_{\text{sup}}$, respectively. We then feed $X_{\text{sup}}$ and the bounding boxes of the human and object into PLM to extract information about the human-object interaction and encode the action purpose. As shown in Figure~\ref{FIG:3}, our network needs to estimate the purpose that the person wants to swing. Subsequently, we feed the feature representation of the action purpose and $\mathcal{X}$ into PTM, which transfers the action purpose to $\mathcal{X}$, enabling the network to learn to attend those objects with that affordance. Finally, the encoded features are fed into CEM, which captures the intrinsic connections between objects with the same affordance and suppresses irrelevant object regions, predicting the common affordance masks.

\subsection{Purpose Learning Module}
As shown in Figure~\ref{FIG:4} (a), we feed the $X_{\text{sup}}$ and the bounding boxes of the person and object into the purpose learning module to estimate the action purpose of the person. We first use the bounding boxes to extract the features of the person and object from $X_{\text{sup}}$ as $X_\text{H}$ and $X_\text{O}$. Subsequently, inspired by human-object interaction and visual relationship detection \cite{zhan2019exploring,zhan2020multi}, the features of the instance (person or object) can provide guidance information about where the network should focus, we use the features of the person and object to activate $X_{\text{sup}}$ respectively to obtain $M_\text{H}$ and $M_\text{O}$: 
\begin{equation}
     M_{\text{O}}=\text{Softmax}(f_{\text{O}}\otimes X_{\text{sup}})\otimes X_{\text{sup}},  \label{eq:no1}
\end{equation}
\begin{equation}
    M_{H}=\text{Softmax}(f_{\text{H}}\otimes X_{\text{sup}})\otimes X_{\text{sup}},  \label{eq:no2}
\end{equation}
where $f_{\text{O}}$ and $f_{\text{H}}$ are the representations of $X_{\text{O}}$ and $X_{\text{H}}$ respectively after global maximum pooling (GMP). $\otimes$ denotes element-wise product. We then use $f_{\text{O}}$ to guide the network to focus on the area of human-object interaction $M_{\text{HO}}$:
\begin{equation}
     M_{\text{HO}}=\text{Conv}(f_{\text{O}}\otimes X_{\text{H}}).  \label{eq:no3}
\end{equation}
Then, $M_{\text{HO}}$ is multiplied with $M_\text{H}$ and $M_\text{O}$ respectively to activate the relevant features of human-object interaction on the global features, which are added together and go through a GMP layer to get the encoding of the action purpose $F_{\text{sup}}$:
\begin{equation}
     F_{\text{sup}}=\text{MaxPooling}((M_{\text{HO}}\odot M_{\text{H}})+(M_{\text{HO}}\odot M_{\text{O}})),  \label{eq:no4}
\end{equation}
where $\odot$ denotes position-wise dot product.
\subsection{Purpose Transfer Module}
After obtaining $F_{\text{sup}}$, we transfer it to each query image to segment the objects that can fulfill that purpose, i.e., having the common affordance. As shown in Figure~\ref{FIG:4} (b), we perform correlation calculations between $F_{\text{sup}}$ and each position of the query image. After normalization, we get the attention probability map that may contain objects with the common affordance. It is used to enhance the feature that may contain the affordance object and suppress the area that does not contain the affordance object. Finally, the attended feature is added back to the original feature to get $\mathcal{X}_\text{T}$:
\begin{equation}
     X_{\text{T}_{i}}=X_i+\text{Softmax}(X_i \otimes F_{\text{sup}})\otimes X_i, i\in [1,n].  \label{eq:no5}
\end{equation}

\subsection{Collaboration Enhancement Module}
There are some inherent characteristics between objects with the same affordance. For example, both cups and bowls have concave areas in the middle so they can be used to hold water. Thus, by discovering intrinsic features from the query image collection \cite{ma2020auto}, objects with the same affordance can be activated and unrelated regions can be suppressed, resulting in better segmentation results (see Figure~\ref{FIG:4} (c)).

\par Inspired by the Expectation-Maximization (E-M) \cite{dempster1977maximum} algorithm and EMANet \cite{li2019expectation}, we run the ``E-step'' and ``M-step'' alternately to obtain a compact set of bases, which are used to reconstruct the feature map of query images. During the iterative process, this set of bases can learn the common properties between the input set of features and thus can segment the common feature attributes. For a set of input feature maps $\mathcal{X}_{\text{T}}=\{X_{\text{T}_1},..., X_{\text{T}_{n}}\}$ ($X_{T_i}\in R^{N\times C}$, $N=W \times H$), they firstly go through a convolution layer to get $\mathcal{F}=\{F_1,..., F_n\}$. Then, a set of base $\mu \in R^{K \times C}$ is initialized. The E-step estimates the latent variable $\mathcal{Z}=\{Z_1,...,Z_{n}\}$, $\mathcal{Z}\in R^{N \times K}$. The weight of the $k$-$th$ basis for the $j$-$th$ pixel of the $i$-$th$ image is calculated as: 
\begin{equation}
     Z_{ijk}=\frac{\kappa(f_{ij},\mu_{k})}{\sum_{l=1}^K \kappa(f_{ij},\mu_{l})},  \label{eq:no6}
\end{equation}
where $f_{ij}$ is the feature of the $j$-$th$ position of the $i$-$th$ image, $\kappa$ denotes the exponential kernel function, i.e., $exp(,)$. Thus, we have $Z_i=Softmax(F_i\mu^T)$. The M-step updates the bases and calculates $\mu$ as the weighted average of $\mathcal{F}$: 
\begin{equation}
    \mu_k=\frac{\sum_{i=1}^n \sum_{j=1}^L z_{ijk}f_{ij}}{\sum_{i=1}^n\sum_{j=1}^L z_{ijk}}.  \label{eq:no7}
\end{equation}
\par After convergence of the E-M iterations, we use $\mu$ and $\mathcal{Z}$ to reconstruct $\mathcal{X}$ to obtain $\tilde{\mathcal{F}}$: $\tilde{F}_i=Z_i\mu$. Finally, $\tilde{\mathcal{F}}$ is mapped to the residue space of $\mathcal{X}$ using convolution and added to $\mathcal{X}$ to obtain $\hat{\mathcal{X}}$, i.e., $\hat{X}_i=X_i+\text{Conv}(\tilde{F}_i$), $\hat{\mathcal{X}}=\{\hat{X}_1,...,\hat{X}_n\}$.

\subsection{Decoder}
For the $i$-$th$ image, the output of the $m-th$ layer is $P_{i}^{m}=\text{Conv}(\text{Unsample}(\text{Conv}(X_{i}^{m})+P_{i}^{m+1})$, $m\in [1,4]$, where $P_{i}^{5}=\text{Conv}(\hat{X_{i}^{5}})$. A convolutional prediction layer is used to get the final output, i.e., $D_{i}^{m}=\text{Conv}(P_{i}^{m})$, $m\in [1,5]$.
\par The cross-entropy loss is used as the training objective of our network. For the prediction $D_i^m$ from the $m$-$th$ layer of the $i$-$th$ image, we calculate the loss $\mathcal{L}_i^m$ as: 
\begin{equation}
\begin{split}
     \mathcal{L}_i^m=-\frac{1}{N}&(\sum_{j\in Y_{+}}log Pr(y_{ij}=1|D_{ij}^m) \\
                         &+\sum_{j\in Y_{-}}log Pr(y_{ij}=0|D_{ij}^m)),  \label{eq:no8}
\end{split}
\end{equation}
where $Pr(y_{ij}=1|D_{ij}^m)$ is the prediction map in which each pixel denotes the affordance confidence. $Y_{+}$ and $Y_{-}$ denote the affordance region pixels set and non-affordance pixels set, $N=H\times W$. The final training objective is defined as:
\begin{equation}
     \mathcal{L}=\sum_{i=1}^{N}\sum_{m=1}^{5}\mathcal{L}_i^m. \label{eq:no9}
\end{equation}

\section{Experiments}
\subsection{Dataset}
\begin{figure}[t]
	\centering
		\includegraphics[scale=.20]{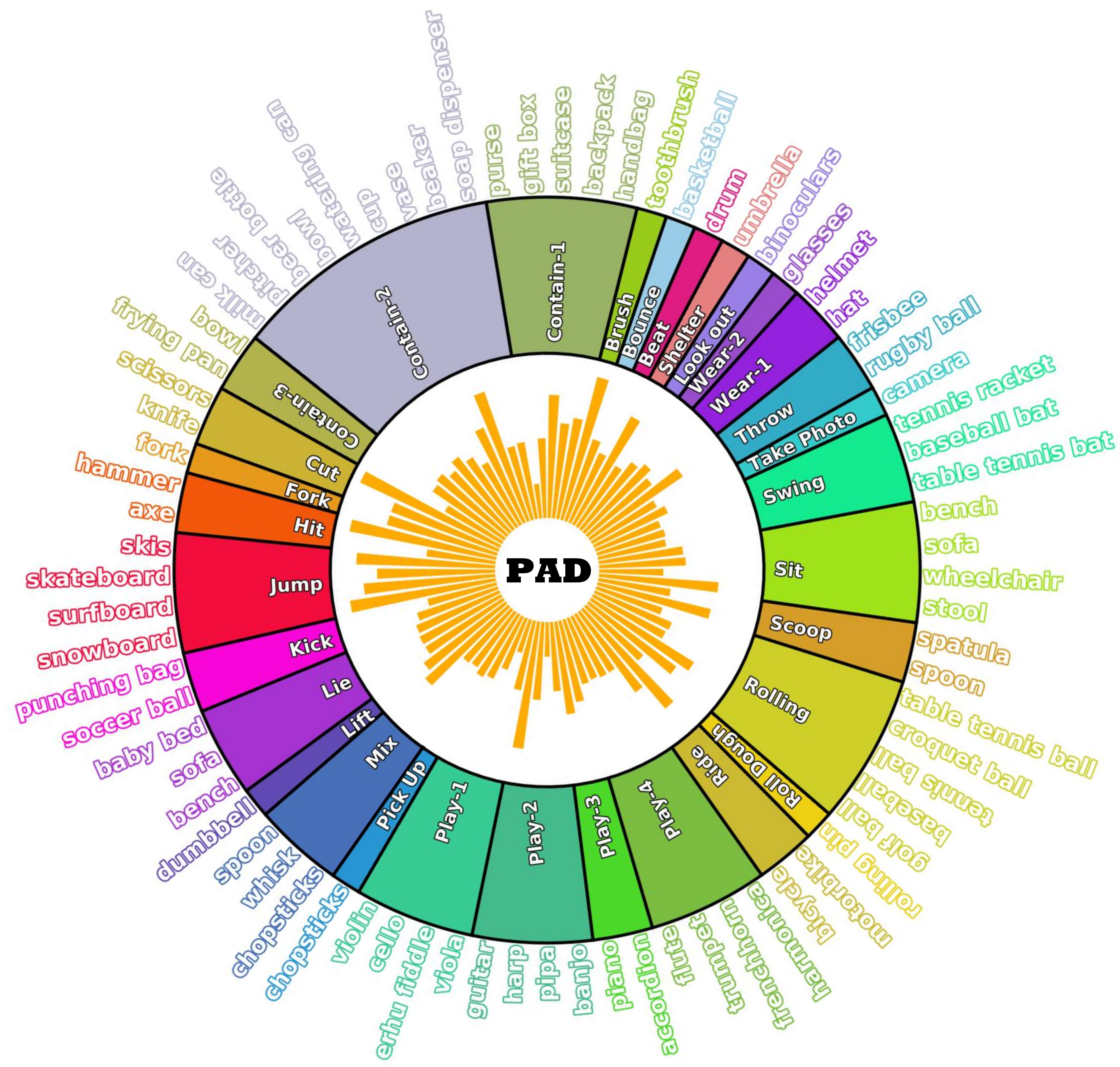}
	\caption{The classification system of the Purpose-driven Affordance Dataset (PAD), which contains 4,002 images covering 72 object classes and 31 affordance classes. See the supplementary material for more details of each affordance category.}
	\label{FIG:5}
\end{figure}

\textbf{Data collection}. We construct the Purpose-driven Affordance Dataset (PAD) with images mainly from ILSVRC \cite{russakovsky2015imagenet}, COCO \cite{lin2014microsoft}, etc. The affordance categories of the dataset is shown in Figure~\ref{FIG:5}. Different images are collected according to the class keywords of the objects to make sure that the images in these datasets cover different scenes and appearance features of the objects, which constitute a challenging benchmark. Additionally, to make the dataset richer in categories, we obtain a part of the images from the Internet to expand the dataset. The description of each affordance and the object categories it contains are provided in the supplementary material. To benchmark different models comprehensively, we follow the k-fold evaluation protocol, where k is 3 in this paper. To this end, the dataset is divided into three parts with non-overlapped categories, where any two of them are used for training while the left part is used for testing. See the supplementary material for more details about the setting. 
\\\textbf{Data annotation}. 1) Category labeling: We create a hierarchy for PAD dataset by selecting 72 common object categories (e.g. ``cups'', ``bowls'', ``basketballs'', etc.) and assigning affordance labels to each object category. An affordance category may contain multiple subcategories, e.g. objects with ``Swing'' affordance include ``tennis rackets'', ``table tennis rackets'', ``golf clubs'', ``baseball bats'', etc., and the appearance feature of these objects varies considerably. An object may contain more than one affordance category, e.g. ``sofa'' and ``bench'' have both ``Sit'' and ``Lie'' affordance labels. 2) Mask labeling: For the images from COCO etc., we select a portion of the mask labels from the original dataset. Since the above dataset may not annotate all the objects belonging to the same affordance, we select these images and annotate them manually. The images downloaded from the web are also manually annotated with the objects containing the affordance category. 3) Purpose image labeling: 656 purpose images are downloaded from the Internet which contains diverse human-object interactions. We annotate the bounding boxes of human and object accordingly.

\begin{table}[t]
\centering
  \footnotesize
  \renewcommand{\arraystretch}{1.}
  \renewcommand{\tabcolsep}{7.5pt}
  \begin{tabular}{r|r||c|c|c||c}
\hline\toprule[1.1pt]
     \rowcolor{mygray}
     & Metrices & $i$ = 1 & $i$ = 2 & $i$ = 3 & Mean $\pm$ Std  \\
   \hline
   \hline
  \multirow{4}{*}{\rotatebox{90}{\textbf{UNet}}} & IoU $\uparrow$ & $.186$  & $.215$ & $.226$ & $.209_{\pm.018}$  \\
   & $E_{\phi}$ $\uparrow$   & $.574$ & $.558$ & $.578$ & $.570_{\pm.009}$     \\  
   & CC $\uparrow$  & $.338$ & $.377$ & $.344$ & $.353_{\pm.017}$              \\
   & MAE $\downarrow$ & $.162$ & $.163$ & $.169$ & $.165_{\pm.003}$       \\      
   \hline
   \multirow{4}{*}{\rotatebox{90}{\textbf{PSPNet}}} & IoU $\uparrow$ & $.261$   & $.244$ & $.295$ & $.267_{\pm.021}$  \\
   & $E_{\phi}$ $\uparrow$    & $.640$ & $.601$  & $.636$ & $.626_{\pm.018}$  \\  
   & CC $\uparrow$     & $.427$   & $.409$ & $.402$ & $.413_{\pm.011}$        \\
   & MAE $\downarrow$  & $.144$  & $.142$  & $.137$  & $.141_{\pm.003}$    \\        
   \hline
   \hline
   \multirow{4}{*}{\rotatebox{90}{\textbf{CPD}}} & IoU $\uparrow$ & $.258$ & $.256$ & $.317$ & $.277_{\pm.028}$   \\
   & $E_{\phi}$ $\uparrow$   & $.615$ & $.601$ & $.630$ & $.615_{\pm.012}$  \\  
   & CC $\uparrow$           & $.413$ & $.386$ & $.433$ & $.411_{\pm.019}$         \\
   &  MAE $\downarrow$  & $.123$ & $.106$ & $.132$ & $.120_{\pm.011}$      \\         
   \hline
   \multirow{4}{*}{\rotatebox{90}{\textbf{BASNet}}} & IoU $\uparrow$ & $.239$ & $.263$  & $.281$ & $.261_{\pm.017}$  \\
   & $E_{\phi}$ $\uparrow$ & $.604$ & $.598$ & $.628$ & $.610_{\pm.013}$      \\  
   & CC $\uparrow$   & $.310$  & $.318$ & $.339$ & $.322_{\pm.012}$               \\
   & MAE $\downarrow$   & $.130$ & $.124$ & $.146$ & $.133_{\pm.009}$   \\ 
   \hline
   \multirow{4}{*}{\rotatebox{90}{\textbf{CSNet}}} & IoU$\uparrow$ & $.173$ & $.210$ & $.238$ & $.207_{\pm.027}$ \\
   & $E_{\phi}$ $\uparrow$ & $.557$ & $.555$ & $.557$ & $.556_{\pm.001}$       \\  
   & CC $\uparrow$  & $.394$  & $.392$ & $.386$ & $.391_{\pm.003}$                 \\
   & MAE $\downarrow$  & $.184$ & $.162$ & $.184$ & $.177_{\pm.010}$     \\        
   \hline
   \hline
   \multirow{4}{*}{\rotatebox{90}{\textbf{CoEGNet}}} & IoU $\uparrow$ & $.281$ & $.262$ & $.289$ & $.277_{\pm.017}$   \\
   & $E_{\phi}$ $\uparrow$    & $.674$ & $.637$  & $.645$  & $.652_{\pm.016}$   \\  
   & CC $\uparrow$   & $.389$ & $.350$  & $.362$  & $.367_{\pm.016}$              \\
   & MAE $\downarrow$ & $.116$ & $.110$ & $.134$ & $.120_{\pm.010}$       \\   
   \hline
   \hline
   \multirow{4}{*}{\rotatebox{90}{\textbf{Ours}}} & IoU $\uparrow$ & $.401$ & $.375$ & $.407$ & $\bm{.394}_{\pm.011}$  \\
   & $E_{\phi}$ $\uparrow$   & $.732$   & $.653$ & $.687$ & $\bm{.691}_{\pm.032}$   \\  
   & CC $\uparrow$        & $.540$    & $.507$ & $.501$ & $\bm{.519}_{\pm.017}$      \\
   & MAE $\downarrow$ & $.103$   & $.116$ & $.122$  & $\bm{.114}_{\pm.008}$ \\
    \hline\bottomrule[1.2pt]
    \end{tabular}
  \caption{Results of different methods on the PAD dataset under the 3-fold test setting. The best results are in \textbf{bold}.}
  \label{Table:1}
  \end{table}

\begin{table}[t]
\centering
  \footnotesize
  \renewcommand{\arraystretch}{1}
  \renewcommand{\tabcolsep}{6pt}
  \begin{tabular}{r||ccc|cc}
\rowcolor{mygray}
\hline\toprule[1.2pt]
                    &  Kick        &  Play-4    &     Contain-2  &        Brush &  Jump                     \\
    \hline\hline
   UNet             &         $.256$       &       $.332$   &          $.253$      &        $.082$   &        $.065$                      \\
   PSPNet          &         $.450$         &   $.452$   &          $.283$           &        $.108$    &        $.122$             \\
   \hline
   CPD              &         $.500$       &     $.505$   &          $.347$         &       $.140$   &        $.162$                   \\
   BASNet            &         $.358$       &       $.408$    &          $.328$      &       $.114$    &        $.109$                    \\
   CSNet             &         $.260$        &       $.393$   &          $.243$         &       $.084$    &        $.043$                   \\
   \hline  
   CoEGNet       &         $.458$         &       $.453$           &          $.321$      &       $.107$  &        $.132$                    \\  
   \hline        
   Ours            &   $\bm{.615}$   &    $\bm{.593}$   &    $\bm{.414}$      &   $\bm{.246}$ &    $\bm{.206}$            \\  
    \hline\bottomrule[1.2pt]
    \end{tabular}
    \caption{Comparison of part results on different affordance categories, using IoU as a metric. Where ``Container-2'' refers to the object with the availability of liquid and ``Play-4'' refers to the objects that a person blows through his mouth to make a sound. See the supplementary material for more details.}
  \label{Table:2}
  \end{table}

\begin{figure*}[t]
	\centering
		\includegraphics[scale=.38]{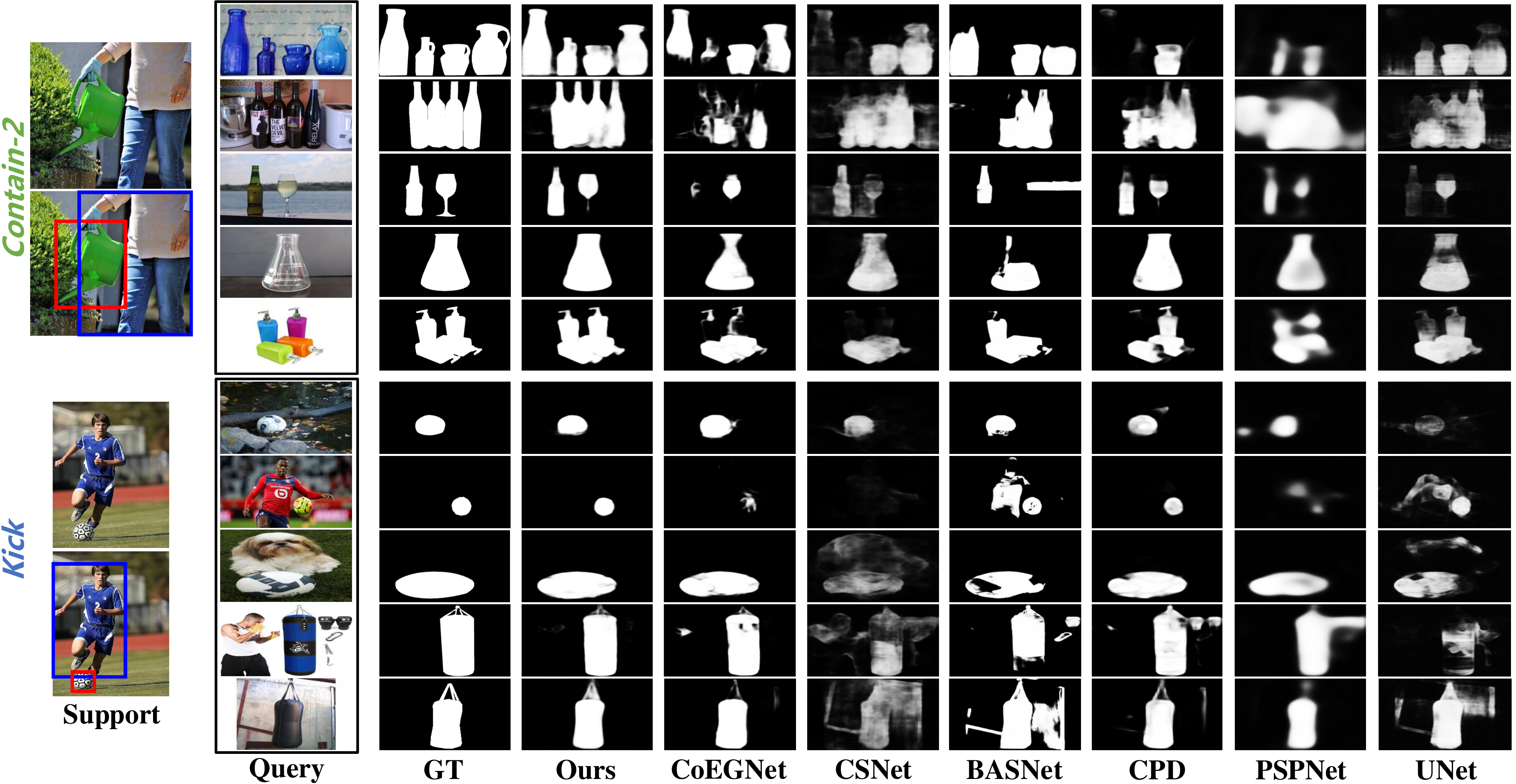}
	\caption{Visual results of different segmentation, saliency detection, co-saliency detection models and our OS-AD on the PAD dataset. OS-AD can learn a better capability to perceive the affordance of objects, i.e., segmenting all objects that complete this purpose and suppressing object regions that are not related to affordance. ``Container-2'' refers to the affordance category that objects can fill liquid.}
	\label{FIG:6}
\end{figure*}

\subsection{Benchmark Setting}
To provide a comprehensive evaluation, four widely used metrics are used to evaluate the performance of affordance segmentation. The \textbf{IoU} metric for segmentation task \cite{long2015fully} is adopted in our task. \textbf{M}ean \textbf{A}bsolute \textbf{E}rror (\textbf{MAE}) \cite{perazzi2012saliency} is used to measure the absolute error between the prediction and ground truth (GT). \textbf{E-measure} ($\bm{E_{\phi}}$) \cite{18IJCAI-Emeasure} is a metric that combines local pixels and image-level average values to jointly capture image-level statistics and local pixel matching information. Pearson \textbf{C}orrelation \textbf{C}oefficient (\textbf{CC}) \cite{le2007predicting} is used to evaluate the correlation between the prediction and GT. We report the average metric score for all test images.

Our method is implemented in Pytorch and trained with the Adam optimizer \cite{kingma2014adam}. The backbone is resnet50 \cite{he2016deep}. The input is randomly clipped from $360\times360$ to $320\times320$ with random horizontal flipping. We train the model for $40$ epochs on a single NVIDIA $1080$ti GPU with an initial learning rate $1e$-$4$. The number of bases in the collaboration enhancement module is set to $K$=$256$. The number of E-M iteration steps is 3. Besides, two segmentation models (\textbf{UNet} \cite{10.1007/978-3-319-24574-4_28}, \textbf{PSPNet} \cite{zhao2017pspnet}), three saliency detection models (\textbf{CPD} \cite{Wu_2019_CVPR}, \textbf{BASNet} \cite{Qin_2019_CVPR}, \textbf{CSNet} \cite{GaoEccv20Sal100K}) and one co-saliency detection models (\textbf{CoEGNet} \cite{deng2020re}) are chosen for comparison.

\subsection{Quantitative and Qualitative Comparisons}

The results are shown in Table \ref{Table:1}, our results outperform all methods on all metrics in the three one-shot learning settings. Measured by the mean IoU value of the three-fold test, our model improves by 42.2\% compared to the co-saliency detection approach. Compared to the best saliency detection model and the best segmentation model, our method improves by 42.2\% and 47.6\% respectively. This indicates that our method can effectively use the action purpose extracted from the support image to guide the segmentation of query images, rather than simply constructing a link between apparent features and affordance. The prediction results generated by each model are shown in Figure~\ref{FIG:6}. It can be seen that our method can detect all the objects with common affordance according to the human purpose. In the 1st and 3rd rows of ``Kick'', the soccer ball and the punching bag are completely different in terms of apparent features, but both are kickable objects. This demonstrates that our method can detect object affordance in the unseen scenarios by extracting the action purpose and collaborative learning strategy.
Meanwhile, we calculate the IoU for each affordance class, and Table~\ref{Table:2} shows the results for several of them. The results from the first three columns show that our method can detect object affordance with different apparent features but belonging to the same affordance, indicating that our model can capture the common relation of the objects in the unseen scenarios well and detect them more completely. The two categories with the worst results out of all the results are ``Jump'' and ``Brush'', respectively. The possible reasons for the poor performance are as follows:  while in the ``Brush'' category, its accompanying actions are not obvious and the size of the objects are relatively small, making it more difficult for the network to extract the features of the objects.
In summary, our approach can learn object affordances well by transferring human purposes to new objects and capturing common features between objects using collaborative learning strategies, with good generalization ability on the task of detecting object affordance in unseen scenarios.

\section{Conclusion}

In this paper, we make the first attempt to deal with a challenging task named one-shot affordance detection, which has practical meaning for real-world applications, such as empowering robots with the ability to perceive unseen affordance. Specifically, we devise a novel one-shot affordance detection (OS-AD) network and construct a benchmark named purpose-driven affordance dataset (PAD). OS-AD exceeds representative models adapted from related areas such as segmentation and saliency detection, which can serve as a strong baseline for this task. PAD contains 4k images with diverse affordance and object categories from different scenes, which can serve as a pioneer test bed for this task. In the future, we plan to deploy the model in a real-world robot to complete some well-defined tasks and evaluate its performance. Going a step further, we attempt to make such a test suite available to the community via a web-based interface.

\section*{Acknowledgments}
\par This work was supported by the National Key R\&D Program of China under Grant 2020AAA0105701, the National Natural Science Foundation of China (NSFC) under Grants 61872327, the Fundamental Research Funds for the Central Universities under Grant WK2380000001.

\bibliographystyle{named}
\bibliography{ijcai21_lhc}

\end{document}